\documentclass{article}

\PassOptionsToPackage{numbers, compress}{natbib}

\usepackage[dblblindworkshop,final]{neurips_2025}

\workshoptitle{UrbanAI}

\usepackage[utf8]{inputenc} 
\usepackage[T1]{fontenc}    
\usepackage{hyperref}       
\usepackage{url}            
\usepackage{booktabs}       
\usepackage{amsfonts}       
\usepackage{nicefrac}       
\usepackage{microtype}      
\usepackage{xcolor}         

\usepackage{amsmath}
\usepackage{subcaption}
\usepackage{graphicx}
\usepackage{float}
\usepackage{rotating}
\usepackage{multirow}

\usepackage{booktabs}

\usepackage{siunitx}
\usepackage{threeparttable}
\usepackage{colortbl}
\usepackage{xcolor}
\definecolor{best}{HTML}{E5FFCC}        
\definecolor{tablerowgray}{gray}{0.97}  

\newcommand{\mrmodel}[1]{%
  \multirow[t]{3}{*}{\parbox[t]{2.6cm}{\raggedright #1}}%
}

\newcommand{\blockgray}{\rowcolor{tablerowgray}} 
\newcommand{\blockwhite}{\rowcolor{white}}       

\title{Real‑time Prediction of Urban Sound Propagation with Conditioned Normalizing Flows}

\author{%
  Achim Eckerle\\
  Stralsund University\\
  \texttt{achim.eckerle@hochschule-stralsund.de } \\
   \And
   Martin Spitznagel \\
   IMLA, Offenburg University \\
   \texttt{martin.spitznagel@hs-offenburg.de} \\
   \AND
   Janis Keuper \\
   IMLA, Offenburg University\\
   \texttt{keuper@imla.ai} \\
}

\begin{document}

\maketitle

\begin{abstract}
Accurate and fast urban noise prediction is pivotal for public health and for regulatory workflows in cities, where the Environmental Noise Directive mandates regular strategic noise maps and action plans, often needed in permission workflows, right-of-way allocation, and construction scheduling. Physics-based solvers are too slow for such time-critical, iterative “what-if” studies. We evaluate conditional Normalizing Flows (Full-Glow) for generating for generating standards-compliant urban sound-pressure maps from 2D urban layouts in real time ($\approx 102~\text{ms}$ per 256$\times$256 map on a single RTX 4090), enabling interactive exploration directly on commodity hardware. On datasets covering Baseline, Diffraction, and Reflection regimes, our model accelerates map generation by \textgreater 2000$\times$ over a reference solver while improving NLoS accuracy by up to 24\% versus prior deep models; in Baseline NLoS we reach 0.65 dB MAE with high structural fidelity. The model reproduces diffraction and interference patterns and supports instant recomputation under source or geometry changes, making it a practical engine for urban planning, compliance mapping, and operations (e.g., temporary road closures, night-work variance assessments).
\end{abstract}

\section{Introduction}

Urban noise is both a public-health and regulatory concern: WHO guidelines link chronic exposure to sleep disturbance and cardiovascular risks, and the EU Environmental Noise Directive mandates recurrent city-scale noise maps and action plans.\citep{WHO2018,END2002} Consequently, urban planners need physically reliable predictions at interactive latencies for permitting and operations. From barrier design to regulating construction projects such as urban drilling, decisions hinge on accurate models of sound propagation in complex cityscapes \cite{Salomons2001, ISO9613}. Although physics-based solvers (ray tracing, FEM) provide high fidelity, their computational cost renders them impractical for the iterative, large-scale “what-if” analyses required in modern city planning \cite{Kinsler2000}.

This computational bottleneck has spurred interest in AI-driven alternatives. Deep learning models, particularly from the U-Net \cite{Ronneberger2015} or GAN \cite{isola2016image} families, can generate sound maps orders of magnitude faster. However, they often trade physical consistency for speed, struggling to accurately model complex wave phenomena like multi-path reflections and diffraction, which are ubiquitous in dense urban canyons \cite{Spitznagel_2025_CVPR}.

We propose leveraging conditional Normalizing Flows (NFs), a class of generative models known for their mathematical rigor and stable training \cite{Dinh2016, Kingma2018}. Their unique invertible architecture allows for exact likelihood computation, making them highly suitable for modeling complex physical distributions \cite{Ardizzone2019}. Specifically, we adopt the Full-Glow architecture \cite{Sorkhei2021FullGlow} to perform an image-to-image transformation from 2D building layouts to sound pressure maps. Our contributions are: (1) a successful application of a fully conditional NF to model distinct urban acoustic phenomena; (2) a quantitative demonstration that our approach significantly outperforms previous deep learning methods in physical accuracy, especially in occluded urban spaces; and (3) validation that NFs can accelerate these simulations by a factor of over 2000 while maintaining high physical fidelity.

\section{Related Work}
\paragraph{Physics-Based Urban Acoustics.}
The gold standard for sound simulation remains physics-based solvers. Geometric acoustics methods like ray-tracing are effective for high-frequency sounds, modeling reflections and shadowing \cite{Salomons2001}. For greater precision, wave-based approaches like the Finite-Element-Method (FEM) solve the underlying wave equations but with a severe computational overhead \cite{Kinsler2000}. Open-source frameworks like NoiseModelling, which implements the CNOSSOS-EU standard, serve as a valuable reference for physically grounded simulations but are too slow for large-scale generative tasks \cite{noisemodelling2025}.

\paragraph{Deep Learning for Physics Simulation.}
AI models have emerged as powerful accelerators. U-Net architectures \cite{Ronneberger2015} are a common baseline for image-to-image tasks but can produce blurry or physically inconsistent results. Generative Adversarial Networks (GANs), such as pix2pix \cite{isola2016image}, can generate sharp, realistic outputs but often suffer from training instability and mode collapse \cite{Creswell2018}. Denoising Diffusion Models (DDPMs) produce high-quality samples but their iterative inference process is computationally intensive, limiting their utility in time-sensitive applications \cite{Ho2020DDPM}.In the urban-acoustics setting, PhysicsGen\citep{Spitznagel_2025_CVPR} provides benchmarked deep baselines on the same dataset, which we use for comparison alongside the public benchmark results\citep{UrbanSoundData2025}.

\paragraph{Normalizing Flows.}
NFs provide a compelling alternative by modeling probability densities explicitly through a series of invertible transformations \cite{Dinh2016}. This allows for stable maximum-likelihood training and exact inference. The Glow model  \cite{Kingma2018Glow} introduced key architectural innovations like invertible 1x1 convolutions, making NFs practical for high-resolution images. The Full-Glow model \cite{Sorkhei2021FullGlow} advances this by conditioning every transformation layer on an input, making it exceptionally well-suited for image-to-image tasks where strong structural guidance is needed. This deep conditioning is what we leverage to enforce physical constraints in the generation of urban sound maps.

\section{Method}
\paragraph{Data.}
We use the \textit{Urban Sound Data} benchmark\citep{UrbanSoundData2025,spitznagel2024urban} as our data source, comprising \textbf{25{,}000} paired samples of OSM-based building masks (inputs) and simulated sound-pressure maps (targets) at $256\times256$ resolution. Simulations follow CNOSSOS-compliant settings via NoiseModelling and are provided in three variants: Baseline, Diffraction (edge diffraction at building corners), and Reflection (up to multiple orders), with predefined train/validation/test splits \citep{noisemodelling2025, Salomons2001}. Intensities are normalized to $[0,1]$; conditioning variables (when present) are min–max scaled. Where indicated, we compare against PhysicsGen\citep{Spitznagel_2025_CVPR}.

\begin{figure}
    \centering
    \setlength{\fboxsep}{0pt}

    \begin{subfigure}[b]{0.24\textwidth}
        \centering
        \fbox{\includegraphics[width=\linewidth]{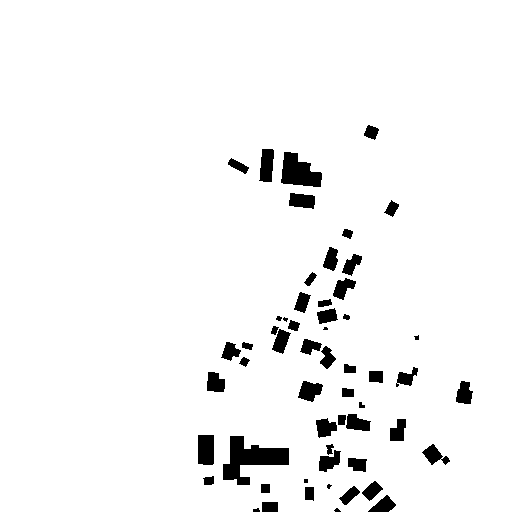}}
        \caption{OSM Layout}
        \label{fig:dataset_osm}
    \end{subfigure}
    \hfill
    \begin{subfigure}[b]{0.24\textwidth}
        \centering
        \fbox{\includegraphics[width=\linewidth]{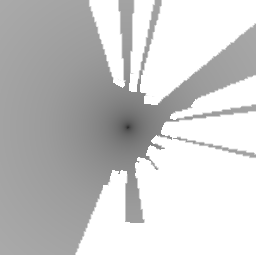}}
        \caption{Baseline Sim.}
        \label{fig:dataset_baseline}
    \end{subfigure}
    \hfill
    \begin{subfigure}[b]{0.24\textwidth}
        \centering
        \fbox{\includegraphics[width=\linewidth]{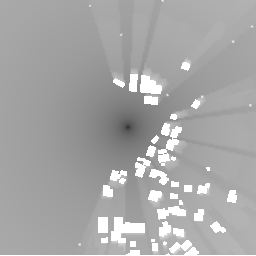}}
        \caption{Diffraction Sim.}
        \label{fig:dataset_diffraction}
    \end{subfigure}
    \hfill
    \begin{subfigure}[b]{0.24\textwidth}
        \centering
        \fbox{\includegraphics[width=\linewidth]{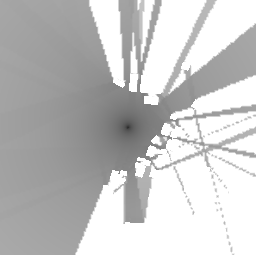}}
        \caption{Reflection Sim.}
        \label{fig:dataset_reflection}
    \end{subfigure}

    \caption{Example data pair: (a) Input urban layout from OSM, and corresponding ground truth simulations for (b) Baseline, (c) Diffraction, and (d) Reflection scenarios.}
    \label{fig:dataset_overview}
\end{figure}

\paragraph{Fully conditioned Glow.}
The architecture extends Glow \citep{Kingma2018} with conditioning injected into \emph{all} invertible steps (ActNorm, invertible $1{\times}1$-convolution, affine coupling), following the Full-Glow design principle \citep{Sorkhei2021FullGlow}. Each flow block receives features from the source pathway (building layout) through a lightweight conditioning network. LU-parameterization keeps $\log|\det(\cdot)|$ tractable in $1{\times}1$-convolutions. Coupling transforms partition channels $(\mathbf{x}_1,\mathbf{x}_2)$ and predict scale $\mathbf{s}$ and translation $\mathbf{t}$ from $\mathbf{x}_1$ plus conditioning $\mathbf{c}$:
\begin{equation}
\mathbf{y}_1 = \mathbf{x}_1, \quad \mathbf{y}_2 = \mathbf{s}(\mathbf{x}_1, \mathbf{c}) \odot \mathbf{x}_2 + \mathbf{t}(\mathbf{x}_1, \mathbf{c})
\end{equation}

\paragraph{Likelihood objective.}
Flows maximize exact data likelihood using the change-of-variables formula
\begin{equation}
\log p(\mathbf{x})=\log p(\mathbf{z})+\sum_{k=1}^{K}\log\bigl|\det\bigl(\frac{\partial h_k}{\partial h_{k-1}}\bigr)\bigr|,
\end{equation}
with standard normal base density $p(\mathbf{z}) = \mathcal{N}(0, I)$ \citep{Rezende2015Variational, Papamakarios2019}. For the conditional mapping $p(\mathbf{x}_{\mathrm{out}} \mid \mathbf{x}_{\mathrm{in}})$, the target flow is conditioned on the source representation (buildings).

\paragraph{Training setup.}
Training was conducted for 1.2M iterations (60 epochs) on 19,908 training samples. Images are processed in a 4-scale multi-scale flow with $[8,8,8,8]$ steps per scale. Batch size is $1$ due to memory constraints ($\approx$14GB VRAM per sample). Adam optimizer ($\beta_1$=0.9, $\beta_2$=0.999) with initial learning rate $10^{-4}$, followed by linear decay to $5{\times}10^{-6}$ from iteration 1M to 1.2M. All experiments performed on NVIDIA RTX 4090 (24GB VRAM), with total training time of $\approx$108 hours per model. Gradient checkpointing reduces peak memory usage by 30-40\%.

\paragraph{Metrics.}
Mean Absolute Error (MAE) and weighted Mean Absolute Percentage Error (wMAPE) are computed separately for Line-of-Sight (LoS) and Non-Line-of-Sight (NLoS) regions, determined via ray-tracing from the central sound source. NLoS regions represent acoustically shadowed areas where direct sound paths are blocked by buildings. wMAPE is computed as $\sum_i |y_i-\hat{y}_i| / \sum_i |y_i|$ with 30 dB threshold to avoid division by near-zero values.

\section{Results}

\begin{table}[H]
\centering
\begin{threeparttable}
\caption{MAE, wMAPE, and runtime across tasks and conditions (smaller is better). Other results are reported from the public benchmark~\citep{UrbanSoundData2025}.}
\label{tab:deeplearning_nlos_acoustic_models}

\begin{tabular}{
  l                                    
  l                                    
  S[table-format=2.2] S[table-format=2.2] 
  S[table-format=2.2] S[table-format=2.2] 
  S[table-format=2.2] S[table-format=2.2] 
}
\toprule
\textbf{Model} & \textbf{Metric}
& \multicolumn{2}{c}{\textbf{Baseline}}
& \multicolumn{2}{c}{\textbf{Reflection}}
& \multicolumn{2}{c}{\textbf{Diffraction}} \\
\cmidrule(lr){3-4}\cmidrule(lr){5-6}\cmidrule(lr){7-8}
& & {LoS} & {NLoS} & {LoS} & {NLoS} & {LoS} & {NLoS} \\
\midrule

\blockgray
\mrmodel{Sim.} & MAE         & 0.00 & 0.00 & 0.00 & 0.00 & 0.00 & 0.00 \\
\blockgray
& wMAPE       & 0.00 & 0.00 & 0.00 & 0.00 & 0.00 & 0.00 \\
\blockgray
& Runtime (ms)
  & \multicolumn{2}{S[table-format=6.0, table-text-alignment=right]}{204700}
  & \multicolumn{2}{S[table-format=6.0, table-text-alignment=right]}{251000}
  & \multicolumn{2}{S[table-format=6.0, table-text-alignment=right]}{206000} \\
\midrule

\blockwhite
\mrmodel{UNet} & MAE         & 2.29 & 1.73 & 2.29 & 5.72 & 0.94 & 3.27 \\
\blockwhite
& wMAPE       & 12.91 & 37.57 & 12.75 & 80.46 & 4.22 & 22.36 \\
\blockwhite
& Runtime (ms)
  & \multicolumn{2}{S[table-format=1.3, table-text-alignment=right]}{0.14}
  & \multicolumn{2}{S[table-format=1.3, table-text-alignment=right]}{0.138}
  & \multicolumn{2}{S[table-format=1.3, table-text-alignment=right]}{0.14} \\
\midrule

\blockgray
\mrmodel{Pix2Pix} & MAE      & \cellcolor{best}\textbf{1.73} & 1.19 & 2.14 & 4.79 & 0.91 & 3.36 \\
\blockgray
& wMAPE    & 9.36 & 6.75 & 11.30 & 30.67 & 3.51 & 18.06 \\
\blockgray
& Runtime (ms)
  & \multicolumn{2}{S[table-format=1.3, table-text-alignment=right]}{0.14}
  & \multicolumn{2}{S[table-format=1.3, table-text-alignment=right]}{0.138}
  & \multicolumn{2}{S[table-format=1.3, table-text-alignment=right]}{0.14} \\
\midrule

\blockwhite
\mrmodel{DDPM} & MAE         & 2.42 & 3.26 & 2.74 & 7.93 & 1.59 & 3.27 \\
\blockwhite
& wMAPE       & 15.57 & 51.08 & 17.85 & 80.38 & 8.25 & 20.30 \\
\blockwhite
& Runtime (ms)
  & \multicolumn{2}{S[table-format=4.2, table-text-alignment=right]}{3986.35}
  & \multicolumn{2}{S[table-format=4.2, table-text-alignment=right]}{3986.35}
  & \multicolumn{2}{S[table-format=4.2, table-text-alignment=right]}{3986.35} \\
\midrule

\blockgray
\mrmodel{Full Glow (ours)} & MAE     & 1.84 & \cellcolor{best}\textbf{0.65} & \cellcolor{best}\textbf{2.06} & \cellcolor{best}\textbf{3.64} & \cellcolor{best}\textbf{0.79} & \cellcolor{best}\textbf{2.63} \\
\blockgray
& wMAPE   & \cellcolor{best}\textbf{8.83} & \cellcolor{best}\textbf{4.52} & \cellcolor{best}\textbf{8.98} & \cellcolor{best}\textbf{22.69} & \cellcolor{best}\textbf{2.43} & \cellcolor{best}\textbf{11.12} \\
\blockgray
& Runtime (ms)
  & \multicolumn{2}{S[table-format=3.2, table-text-alignment=right]}{101.70}
  & \multicolumn{2}{S[table-format=3.2, table-text-alignment=right]}{102.30}
  & \multicolumn{2}{S[table-format=3.2, table-text-alignment=right]}{107.62} \\
\bottomrule
\end{tabular}

\begin{tablenotes}[flushleft]
\footnotesize
\item Best values per column are highlighted in light green. All metrics are averaged over 1,245 test samples for each scenario.
\end{tablenotes}

\end{threeparttable}
\end{table}

\paragraph{Quantitative Analysis.}
Table \ref{tab:deeplearning_nlos_acoustic_models} shows a quantitative comparison against prior deep learning models from \cite{Spitznagel_2025_CVPR} on the same dataset splits. Our model sets a new benchmark in almost all scenarios. For the \textbf{Baseline} condition, it achieves an NLoS-MAE of only \textbf{0.65 dB}, a 45\% improvement over the best competing model (Pix2Pix). This indicates an exceptional ability to model basic acoustic shadowing. In the more complex \textbf{Diffraction} scenario, our model again leads with an NLoS-MAE of \textbf{2.63 dB}. Most notably, in the challenging \textbf{Reflection} scenario, where multi-path interference is key, our model achieves an NLoS-MAE of \textbf{3.64 dB}, a 24\% improvement over Pix2Pix. These strong NLoS results confirm the model's superior ability to capture complex wave phenomena in occluded urban spaces.

\begin{figure}[htbp]
    \centering
    \captionsetup{font=small, justification=centering}

    \textbf{Baseline Scenario} \\
    \includegraphics[width=1\linewidth]{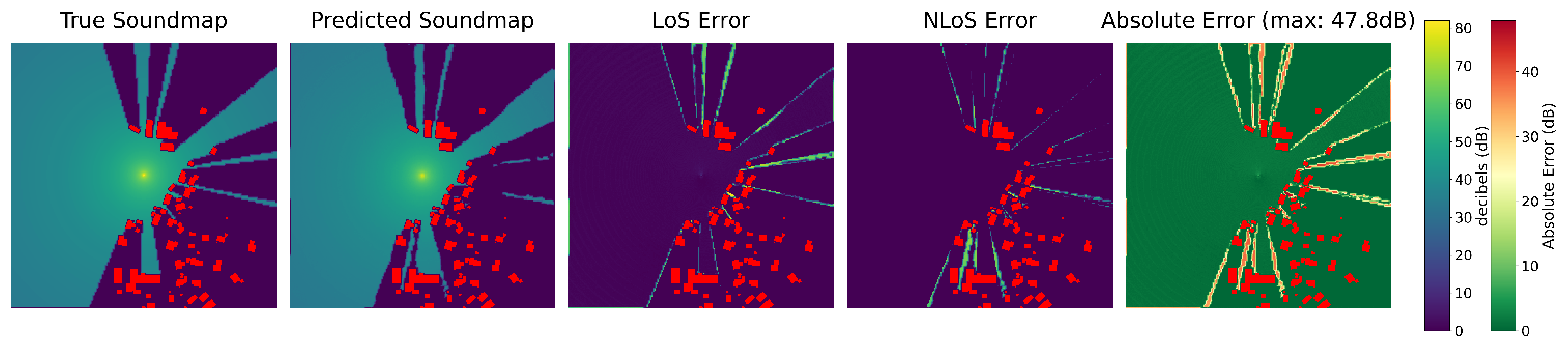}
    \vspace{-4mm} 

    \textbf{Reflection Scenario} \\
    \includegraphics[width=1\linewidth]{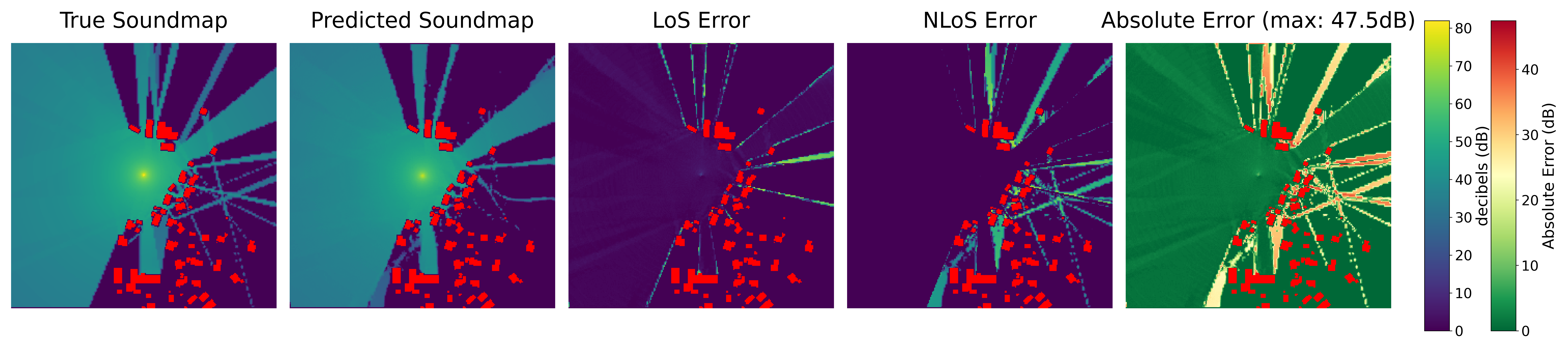}
    \vspace{-4mm} 

    \textbf{Diffraction Scenario} \\
    \includegraphics[width=1\linewidth]{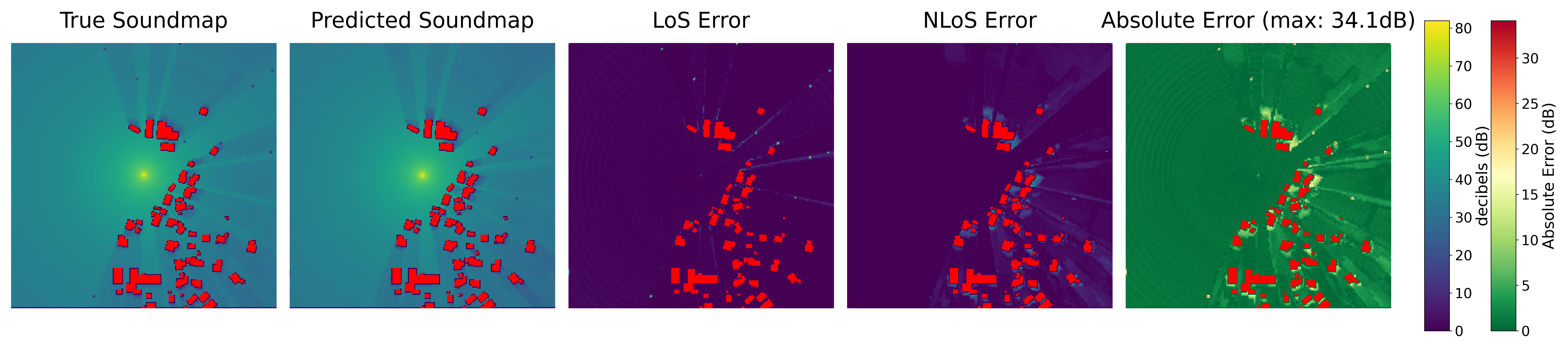}

    \caption{Visual comparison of our model's predictions (center column) against the ground truth (left column) for the Baseline, Reflection, and Diffraction scenarios. The absolute error maps (right column) confirm high physical fidelity across all cases.}
    \label{fig:qualitative_comparison}
\end{figure}

\paragraph{Qualitative and Structural Analysis.}
As shown in Figure \ref{fig:qualitative_comparison}, the sound maps generated by Full-Glow are visually almost indistinguishable from the ground truth simulations. The model correctly reproduces the sharp acoustic shadows in the Baseline case, the characteristic soft-edged fans of diffraction, and the complex interference patterns in the Reflection scenario. The absolute error maps confirm that errors are small and localized, avoiding the systematic blurring seen in other models. The high structural similarity is further confirmed by SSIM scores, with mean values of \textbf{0.92} for Baseline, \textbf{0.96} for Diffraction, and \textbf{0.85} for Reflection, indicating excellent preservation of the sound field's spatial structure.

\begin{figure}[htbp]
    \centering
    \textbf{Model comparison with 95\% confidence interval} \\
    \includegraphics[width=1\linewidth]{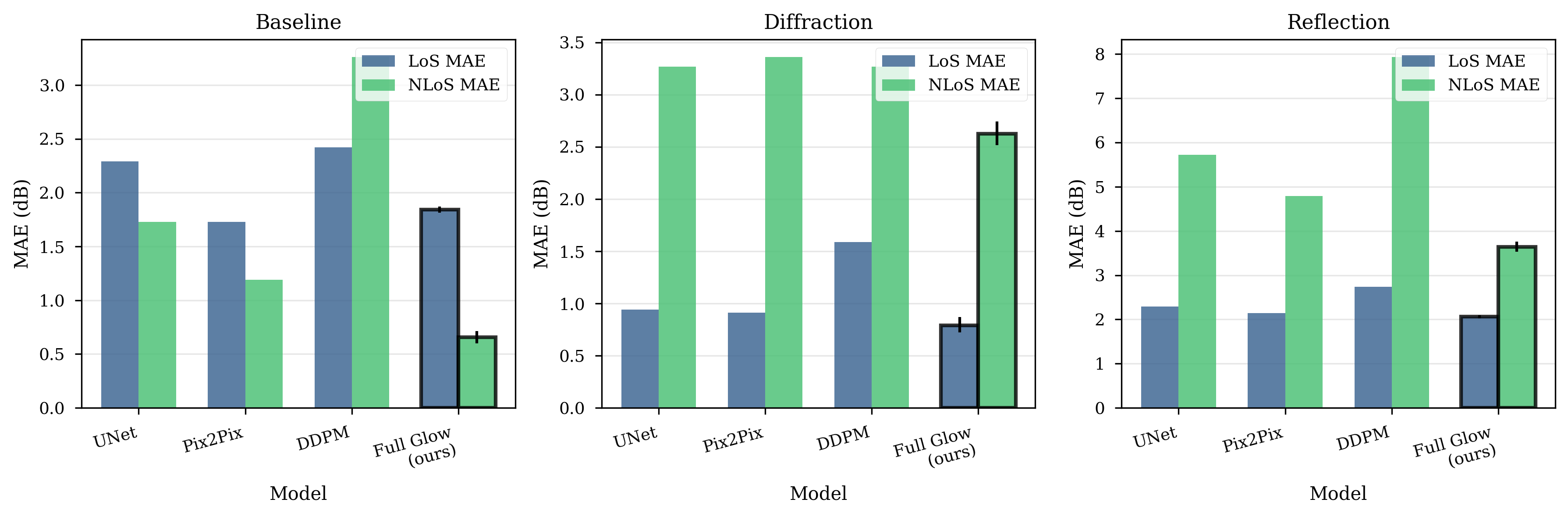}
    \caption{Model comparison with 95\% confidence intervals across all three acoustic scenarios, computed over 1,245 test samples per scenario. The non-overlapping confidence intervals confirm the statistical significance of our Full-Glow model's performance gains, particularly in acoustically shadowed (NLoS) regions.}
    \label{fig:model_comparison}
\end{figure}

\paragraph{Statistical Significance.}
Figure~\ref{fig:model_comparison} extends our quantitative analysis by providing 95\% confidence intervals across all test samples. The narrow confidence intervals around our Full-Glow model's performance demonstrate that the improvements reported in Table~\ref{tab:deeplearning_nlos_acoustic_models} are statistically robust and not driven by outliers. Notably, the confidence intervals for NLoS errors do not overlap between our method and competing approaches in any scenario, confirming statistical significance. The consistently larger confidence intervals in the Reflection scenario across all models reflect the inherent stochasticity of multi-path interference patterns, yet our approach maintains the tightest bounds even in this challenging regime.

\section{Conclusion}
This paper reports a fully conditioned normalizing-flow approach for urban sound propagation. On three scenarios (Baseline, Diffraction, Reflection), the method achieves accurate LoS/NLoS metrics and while providing large inference-time speedups over classical simulation. They offer a compelling balance of generative speed, model stability, and physical accuracy, making them a highly promising tool for practical applications in urban planning, noise assessment, and beyond. Limitations include sensitivity to multi-effect complexity (reflections remain hardest) and high training memory.

\section*{Funding Acknowledgement}
The authors acknowledge the financial support by the German Federal 
Ministry of Education and Research (BMBF) in the program “Forschung an Fachhochschulen in Kooperation mit Unternehmen (FH-Kooperativ)” within the joint project "KI-Bohrer" under 
grant 13FH525KX1\\ \url{https://www.ki-bohrer.de/}.
\newpage

\small{
\bibliographystyle{unsrtnat}
\bibliography{refs}
}

\end{document}